%% file: main_aaai.tex
\documentclass[letterpaper]{article} 
\usepackage{times}  
\usepackage{helvet}  
\usepackage{courier}  
\usepackage[hyphens]{url}  
\usepackage{graphicx} 
\usepackage{natbib}  
\usepackage{caption} 
\usepackage{algorithm}
\usepackage{algorithmic}

\usepackage{newfloat}
\usepackage{listings}
\DeclareCaptionStyle{ruled}{labelfont=normalfont,labelsep=colon,strut=off} 
\floatstyle{ruled}
\newfloat{listing}{tb}{lst}{}
\floatname{listing}{Listing}
\usepackage{booktabs}       
\usepackage{graphicx, caption}
\usepackage{multirow}
\usepackage{makecell}
\usepackage{subfig}
\input{math_commands.tex}

\newcommand{\method}{Redactor}
\newcommand{\model}[1]{{\em #1}}

\newboolean{techreport}
\setboolean{techreport}{true}

\ifthenelse{\boolean{techreport}}
{\usepackage{techreport}}
{\usepackage[]{aaai23}}

\title{\method{}: A Data-centric and Individualized Defense Against Inference Attacks}
\author{
    Geon Heo,
    Steven Euijong Whang
}
\affiliations{
    KAIST\\

    \{geon.heo, swhang\}@kaist.ac.kr
}

\usepackage{bibentry}

\begin{document}

\maketitle

\begin{abstract}
Information leakage is becoming a critical problem as various information becomes publicly available by mistake, and machine learning models train on that data to provide services. As a result, one's private information could easily be memorized by such trained models. Unfortunately, deleting information is out of the question as the data is already exposed to the Web or third-party platforms. Moreover, we cannot necessarily control the labeling process and the model trainings by other parties either. In this setting, we study the problem of {\em targeted disinformation generation} where the goal is to dilute the data and thus make a model safer and more robust against inference attacks on a specific target (e.g., a person's profile) by only inserting new data. Our method finds the closest points to the target in the input space that will be labeled as a different class. Since we cannot control the labeling process, we instead conservatively estimate the labels probabilistically by combining decision boundaries of multiple classifiers using data programming techniques. Our experiments show that a probabilistic decision boundary can be a good proxy for labelers, and that our approach is effective in defending against inference attacks and can scale to large data.
\end{abstract}

\section{Introduction}

Information leakage is becoming a serious problem as personal data is being used to train machine learning (ML) models. Personal data can be leaked through AI chatbots\,\cite{iruda} and the Web\,\cite{flickr} among others. Furthermore, there are various privacy threats on ML models including inference attacks \cite{shokri2017membership, hayes2019logan,choo2020label} and reconstruction attacks \cite{fredrikson2015model}. Defending against such leakage is critical for safe and robust AI.

\begin{figure}[t]
\centering
    \subfloat[Conventional Inference Attack Defense]{
        {\includegraphics[scale=0.465]{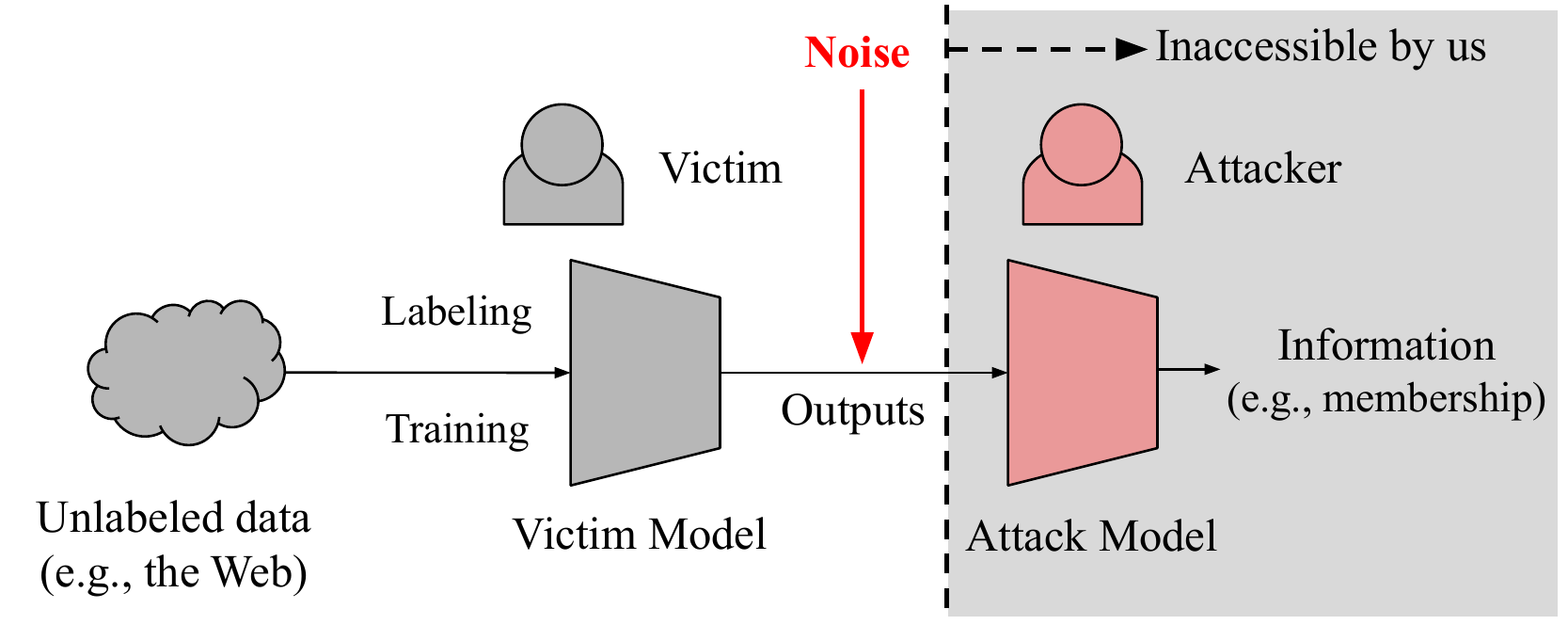}}
        }\hfill
    \subfloat[\method{}]{
        {\includegraphics[scale=0.465]{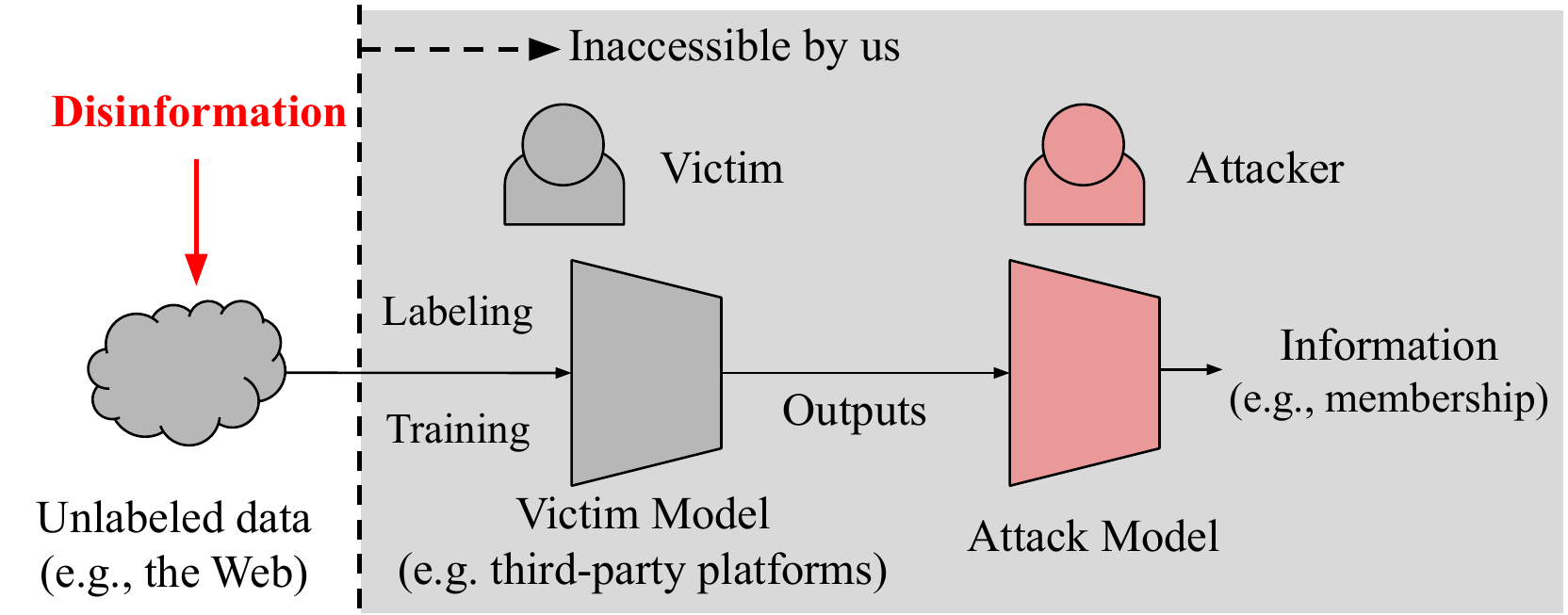}}
        }
    \caption{(a) Existing inference attack defenses (e.g., add noise to model's output) are not feasible when the victim models are owned by third parties. (b) Instead, we assume the realistic setting where we can only add disinformation to the unlabeled training data, which is presumably labeled and used for model training by unknown model owners.}
\label{fig:setting}
\end{figure}

In many cases, it is impossible to delete one's information that is published on the Web or uploaded on a third-party platform. Even if the original data is deleted by request, there is no way to prevent someone from extracting that information elsewhere by attacking the model of the unknown third-party platform. Moreover, there is also no control over the model training process where anyone can train a model on the publicized data. Therefore, conventional privacy techniques or defenses that require ownership of the data or model cannot be used here.

The only solution is to take a data-centric approach and add new data that ``dilutes'' an individual's personal information, which we refer to as {\em disinformation}. An analogy is blacking out or redacting text where the reader knows there is some information, but cannot read it. We thus define the problem of {\em targeted disinformation generation} where the goal is to generate disinformation that indirectly makes a victim model less likely to leak personal information to an attack model without any access to the victim model (Figure~\ref{fig:setting}). We assume the disinformation will be eventually picked up automatically by crawlers for model training, which is a common assumption in the AI Security literature\,\cite{DBLP:conf/nips/ShafahiHNSSDG18,DBLP:conf/uss/SuciuMKDD18,DBLP:conf/ijcai/ChenJLPSS19}. From an ethical perspective, our disinformation is intended to protect one's information from inference attacks.

Our solution is motivated by clean-label targeted poisoning methods\,\cite{DBLP:conf/uss/SuciuMKDD18,DBLP:conf/nips/ShafahiHNSSDG18,DBLP:conf/icml/ZhuHLTSG19,DBLP:journals/titb/KermaniSRJ15} that degrade the model performance only on a target example. We would like to utilize such techniques to change the output of the unknown models (e.g., third-party platforms) and protect target examples from indirect privacy attacks. However, many of these poisoning techniques implicitly rely on a transfer learning\,\cite{DBLP:journals/tkde/PanY10} setup where a pre-trained model is available and exploit the fact that the feature space is fixed for optimizing the poisoning. While transfer learning benefits certain applications (e.g., NLP or vision tasks), it is not always available, especially for structured data where there is no efficient and generally-accepted practice\,\cite{borisov2021deep}. However, structured data is key to our problem as most personal information is stored in this format (e.g., name, gender, age, race, address, and more). In order to primarily support structured data, we thus need to assume {\em end-to-end training} where we cannot count on a fixed feature space. Although existing techniques have also been extended for end-to-end training, we show their performances are not sufficient.

\begin{figure}[t]
\centering
    \includegraphics[width=0.9\columnwidth]{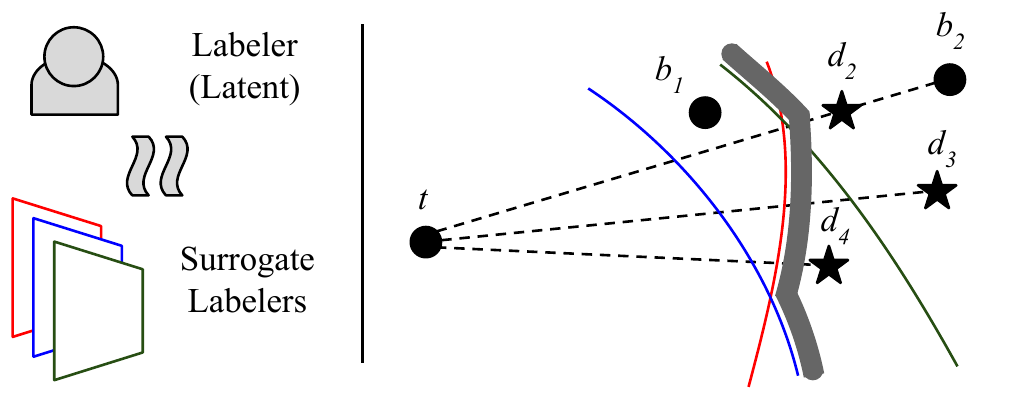}
\caption{A labeler is approximated as a probabilistic model that combines surrogate labelers. The decision boundaries of surrogate labelers are shown as thin lines while the probabilistic decision boundary (PDB) is the thick gray line. Using the base example $b_2$, \method{} can generate the disinformation example $d_2$ that is close to $t$, but still labeled differently. In addition, \method{} can generate other realistic points $d_3$ and $d_4$ using a generative model where $d_4$ happens to be even closer to $t$.}
\label{fig:approach}
\end{figure}

Since we cannot rely on a fixed feature space, we instead utilize the input space to find the best disinformation that is close to the target, but labeled differently. How do we know the true label of the disinformation? Since we do not have access to the labelers, one of our key contributions is a novel adaptation of data programming\,\cite{DBLP:journals/vldb/RatnerBEFWR20,DBLP:journals/pvldb/RatnerBEFWR17,DBLP:conf/sigmod/RatnerBER17} to conservatively estimate human behavior using a {\em probabilistic decision boundary} (PDB) produced by combining multiple possible classifiers. In our setting, we make the generative model produce the probability of an example having a class that is different than the target's class. By limiting this probability to be above a tolerance threshold, we now have a conservative decision boundary. This approach is agnostic to the victim model. We call our system \method{}, and Figure~\ref{fig:approach} illustrates our overall approach.

Our contributions: (i) We define the {\em targeted disinformation generation} problem.  (ii) We suggest a novel data-centric and model-agnostic defense using data programming and generative models without any access to the training process of victim models. (iii) We empirically demonstrate that our solution is effective in defending against membership inference attacks and scales to large data. 



\section{Background}
\label{sec:background}

\subsection{Membership Inference Attack (MIA)}

Among various types of adversarial attacks, exploratory attacks are used to extract information from models. The dominant attack most related to our work is the membership inference attack (MIA). The goal of an MIA is to train an attack model that predicts if a specific example was used to train a victim model based on its confidence scores and loss values. Formally, an attacker trains an attack model $A$ satisfying $A:s\rightarrow\{0,1\}$ where the input $s$ is the confidence score or loss value of the victim model $V$ for an example $x$, and 1 means that $x$ is a member of the training set of $V$. 

Many defenses\,\cite{DBLP:conf/ccs/JiaSBZG19,DBLP:conf/codaspy/LiL021,DBLP:conf/ndss/Salem0HBF019} have been proposed against MIAs, but most of them assume that accessing the victim model $V$ is possible. For example, MemGuard\,\cite{DBLP:conf/ccs/JiaSBZG19} is a state-of-the-art defense that adds noise to the model's output to drop the attack model's performance. Other techniques include adding a regularizer to the model's loss function\,\cite{DBLP:conf/codaspy/LiL021} and applying dropout or model stacking techniques\,\cite{DBLP:conf/ndss/Salem0HBF019}. 

Such model modifications are not possible in our setting where we assume no access to the model. We thus design a new approach using a targeted poisoning objective to indirectly change the victim model's performance on the target (confidence score and loss value).

\subsection{Targeted Poisoning}

Targeted poisoning attacks have the goal of flipping the predictions on specific targets to a certain class. A na\"ive approach is to add examples that are identical to the target, but with different labels. Unfortunately, such an approach would only work if one has complete control over the labeling process, which is unrealistic. Instead, the poison $p$  needs to be different enough from the target to be labeled differently by any human. Yet, we also want $p$ to be close to the target. 

\begin{figure}[t]
\centering
    \subfloat[Transfer Learning Scenario]{
        {\includegraphics[scale=0.84]{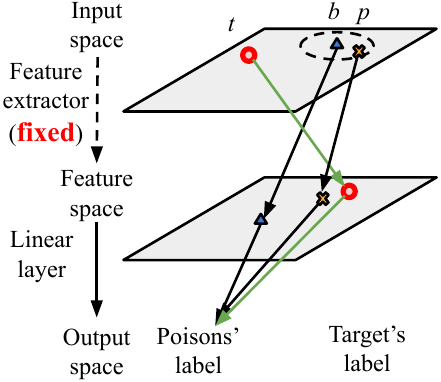}}
        \label{fig:scenario1}
        }
    \hspace{0.1cm}
    \subfloat[End-to-end Training Scenario]{
        {\includegraphics[scale=0.87]{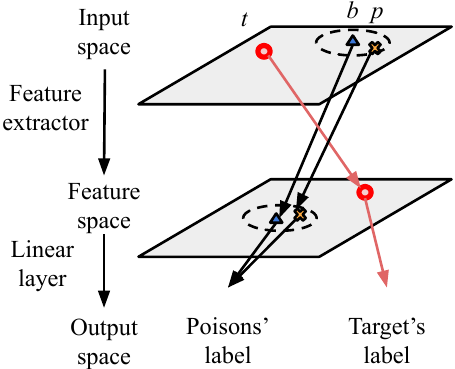}}
        \label{fig:scenario2}
        }
    \caption{In a transfer learning scenario (a), the feature space is fixed, making it possible to optimize on both the input and feature spaces. In an end-to-end scenario (b), however, the feature space may change after the model trains, so optimizing on the feature space may not be effective.}
\label{fig:scenarios}
\end{figure}

The state-of-the-art targeted poisoning attacks include Convex Polytope Attack (CPA)\,\cite{DBLP:conf/icml/ZhuHLTSG19} and its predecessors\,\cite{DBLP:journals/corr/abs-1712-05526,DBLP:conf/uss/SuciuMKDD18,DBLP:conf/nips/ShafahiHNSSDG18}, which also do not assume any control over the labeling and generate poison examples that are similar to the base examples, but have the same predictions as the target. These techniques are not involved in the model training itself, but generate poisons that are presumably added to the training set. The goal is to generate examples close to the target in the feature space while being close to a base example in the input space. To find an optimal poison satisfying such conditions, CPA utilizes a fixed feature extractor, which is effective when the victim uses transfer learning (see Figure~\ref{fig:scenario1}). 

In end-to-end training, however, all the layers of the model are trainable where any feature space that is not the input space may change after model training with the poison. Therefore, CPA's optimization may not be effective because any distance on the feature space corresponding to each layer can change arbitrarily. Figure~\ref{fig:scenario2} illustrates this point where the poison example $p$ can still be close to the base example $b$ on a feature space that is not the input space even after CPA's optimization. Empirical results for this analysis can be found in \ifthenelse{\boolean{techreport}}{the appendix.}{our technical report\,\shortcite{redactortr}.} Although \cite{DBLP:conf/icml/ZhuHLTSG19} suggests the extension of applying CPA on every layer of the network for end-to-end training, it is inefficient and does not fundamentally solve the problem. We thus need a completely different solution that does not utilize the feature space for optimization.


\section{Methodology: \method{}}
\label{sec:redactor}

We design an optimization problem of generating targeted disinformation for end-to-end training based on the targeted poisoning objective. We describe our objectives and introduce the overall process of \method{}. In end-to-end training, we can only utilize the input space and need to generate a disinformation that is as close as possible to the target example, but likely to be labeled as a different class from the target. Suppose that a human labeler has a mental decision boundary for labeling. In order to satisfy both conditions, the disinformation must be the closest point on the other side based on this decision boundary as we define below: 
\begin{equation}
\label{eq:problem1}
\begin{split} 
\min_{\{d_j\}} \; \sum_{j=1}^{N_d} dist(d_j, t) \\
\textrm{s.t. }\; HumanLabel(\,d_j) \neq c_t\\
d_j \in C_{real},\: \forall j \in [1 \ldots N_d]
\end{split}
\end{equation}
where $t\in\mathbb{R}^D$ is the target example, $d_j\in\mathbb{R}^D$ is the $j$th disinformation among a budget of $N_d$ disinformations, $c_t$ is $t$'s class, and $C_{real}\subseteq\mathbb{R}^D$ is a set that conceptually contains all possible realistic candidates where $D$ is the number of features.
However, since we do not have control of the labeling and thus do not know the decision boundary, we propose to use surrogate classifier models as a proxy for human labeling, which we call {\em surrogate labelers}. This approach is inspired by ensemble techniques commonly used for forging black-box attacks\,\cite{DBLP:conf/icml/ZhuHLTSG19,DBLP:conf/iclr/LiuCLS17}. We do not assume that the surrogate labelers are highly accurate. However, when combining these models, we assume that we can find a {\em conservative decision boundary} that can confidently tell whether an example will be labeled differently than the target. 
Based on Equation~\ref{eq:problem1}, we now formulate our optimization problem as follows:
\begin{equation}
\label{eq:problem2}
\begin{split} 
\min_{\{d_j\}} \; \sum_{j=1}^{N_d} ||d_j - t||^2 \\
\textrm{s.t. }\; \argmax_c \; M_c(\phi,\;d_j) \neq c_t\\
\max_{c \neq c_t} \; M_c(\phi,\;d_j) \geq \alpha \\
d_j \in C_{cand},\: \forall j \in [1 \ldots N_d]
\end{split}
\end{equation}
where $M_c(\phi,\;x)$ is the probabilistic generative model that combines surrogate labelers $\phi$ and returns the probability of an example $x$ being in class $c$, $C_{cand}$ is a realistic candidate set that we generate, and $\alpha$ is the tolerance threshold for $M_c$. We use common pre-processings where numeric features are normalized, and categorical features are converted to have numerical values using one-hot encoding.

\begin{figure}[t]
\centering
    \includegraphics[width=\columnwidth]{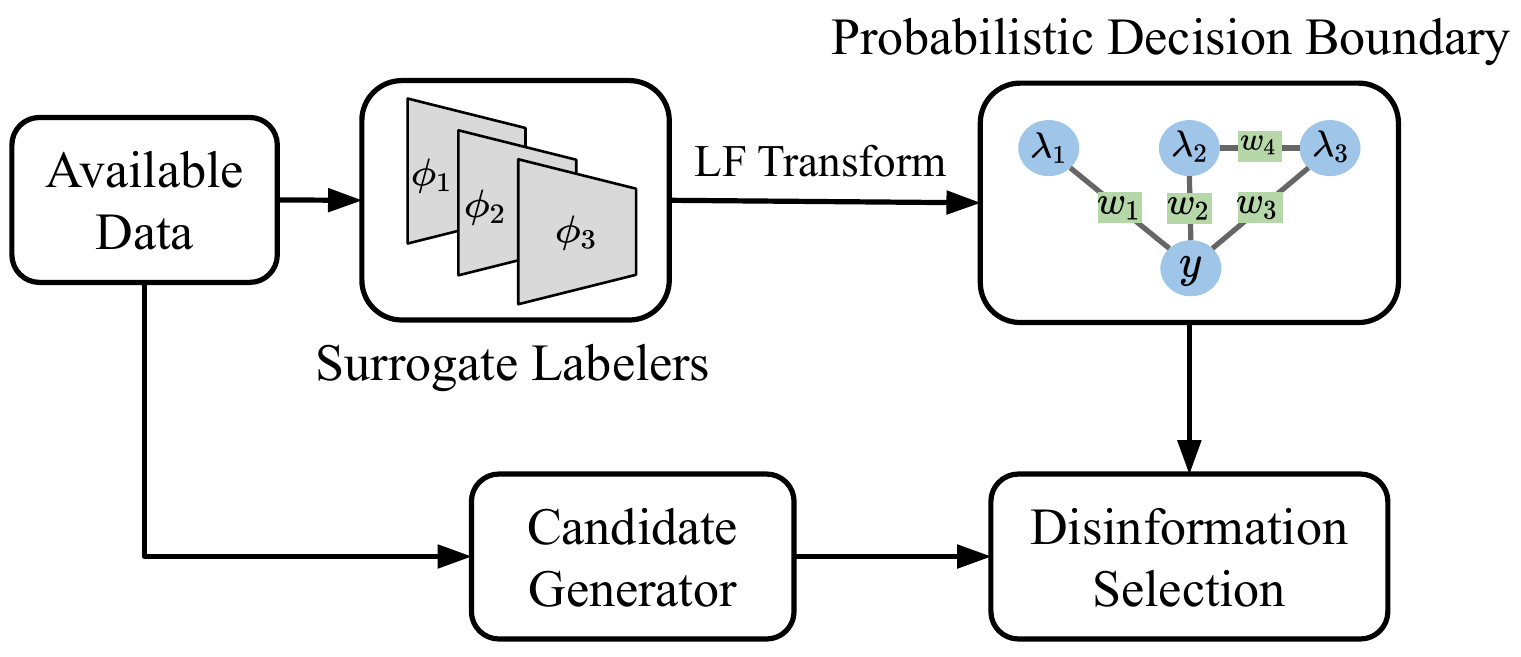}
\caption{\method{} runs in four stages: surrogate labeler training, Probabilistic Decision Boundary (PDB) creation, candidate generation, and disinformation selection.}
\label{fig:method}
\end{figure}

\method{} generates disinformation in four stages (Figure~\ref{fig:method}): training surrogate labelers on the available data, creating a PDB, generating realistic candidates, and selecting the examples that will be used as disinformation. In the next sections, we cover each component in more detail. The overall algorithm is in \ifthenelse{\boolean{techreport}}{the appendix}{our technical report\,\shortcite{redactortr}}.

\subsection{Training Surrogate Labelers}
\label{sec:trainingsurrogatemodels}

When choosing surrogate labelers, it is useful to have a variety of models that can complement each other in terms of performance as they are not necessarily highly accurate. Similar strategies are used in data programming and ensemble learning~\cite{bagging,boosting}. However, our goal is not necessarily improving the overall accuracy of the combined model, but ensuring a conservative PDB. That is, there should be few false positives where a disinformation that is predicted to be on the other side of the target is actually labeled the same. 

Another issue is that we may only have partial data for training surrogate labelers because the data is too large or unavailable. Indeed, if we are protecting personal information on the Web, it is infeasible to train a model on the entire Web data. However, we argue that we only need data that is in the vicinity of the target and contains some examples in different classes as well. We only require that the PDB approximates the decision making around the target. In our experiments, we show how \method{} can scale to large data by properly selecting partial data.

\subsection{Probabilistic Decision Boundary (PDB)}
\label{sec:probabilisticdecisionboundaries}

We now explain how to generate a conservative PDB for identifying examples that will very likely not be labeled the same as the target. We utilize multiple surrogate labelers and combine them into a single probabilistic model using data programming\,\cite{DBLP:journals/vldb/RatnerBEFWR20,DBLP:journals/pvldb/RatnerBEFWR17,DBLP:conf/sigmod/RatnerBER17} techniques, which combines multiple labeling functions into a label model that produces probabilistic labels. 

The data programming framework assumes that each labeling function (LF) can output one of the classes as a prediction or an abstained prediction (-1) if not confident enough. The abstained prediction is necessary for making the LFs as reliable as possible. Thus, we transform each surrogate labeler $\phi_i$ as follows:
\[
\lambda(\phi_i,\;x) = \left\{
                \begin{array}{ll}
                  -1 (Abstain) & \max_c \phi^{(c)}_i(x) \leq \beta/C \\
                  \argmax_c \phi^{(c)}_i(x) & \text{otherwise} \\
                \end{array}
              \right.
\]
where $\beta$ is used to determine when to abstain, $c\in [1 \ldots C]$ is a class, and $\phi_i$ outputs a $C$-dimensional probability vector.

We train a probabilistic generative model $M$ with latent true labels $Y$ using the label matrix $\Lambda_{\phi,x}$ where $\Lambda^{(i,j)}_{\phi,x}=\lambda(\phi_i,\;x_j)$:
\begin{align*}
&P_{w}(\Lambda_\phi,\;Y) = Z_w^{-1}\:\exp\left(\sum_{k=1}^l w^T\; Corr_k(\Lambda_\phi,\;y_k)\right) \\
&\hat{w}=\argmax_w \: \log \sum_Y P_{w}(\Lambda_\phi,\;Y)\\
&M(\phi,\;d) = P_{\hat{w}}(Y|\;\Lambda_{\phi,d}).
\end{align*}
Here $Corr$ values are binary values indicating all possible correlations between LFs and the latent $Y$, $Z_w^{-1}$ is the constant for normalization, and $w$ has the weights of the generative model corresponding to each correlation. 

We then use $M$ with $\phi$ as the PDB. For each example $d$, $M$ returns probability values for each class. Then $d$ is considered to be in a different class than the target $t$ if the class with the maximum probability is not $t$'s class, and the maximum probability is at least the tolerance threshold $\alpha$.

\subsection{Candidate Generation \& Disinformation Selection}
\label{sec:disinformationgeneration}

Given a target, we would like to find the closest possible points that would be labeled differently. Obviously we cannot use the target itself as it would not be labeled differently. Instead, we utilize the PDB to find the closest point beyond the projected real decision boundary. We use watermarking\,\cite{quiring2018forgotten,DBLP:journals/corr/abs-1712-05526,hitaj2018have,DBLP:conf/nips/ShafahiHNSSDG18} techniques where a watermark of the target is added to the base example to generate disinformation using linear interpolations. While this approach works naturally for image data (i.e., the disinformation image is the same as the base image, but has a glimpse of the target image overlaid), structured data consists of numeric, discrete, and categorical features, so we need to perform watermarking differently. For numeric features, we can take linear interpolations. For discrete features that say require integer values, we use rounding to avoid outputting real numbers as a result of the interpolation. For categorical features, we choose the base's value or target's value, whichever is closer. More formally:
\begin{align*}
&\text{Numeric}:\:d^{(i)}=\gamma t^{(i)}+(1-\gamma) b^{(i)} \\
&\text{Discrete}:\:d^{(i)}=round(\gamma t^{(i)}+(1-\gamma) b^{(i)})\\
&\text{Categorical}:\:d^{(i)}=round(\gamma) t^{(i)} +round(1-\gamma)b^{(i)}
\end{align*}
where $d$ is the disinformation example, $t$ is the target, $b$ is a base example, $x^{(i)}$ is $x$'s attributes corresponding to the feature index set $i$, $round(x)=\left \lfloor x+0.5 \right \rfloor$, and $0\leq\gamma\leq1$.

In order to increase our chances of finding disinformation closer to the target, we can use GANs to generate more bases that are realistic and close to the decision boundary. Among possible GAN techniques for tabular data~\cite{DBLP:journals/corr/abs-1911-03274, DBLP:conf/mlhc/ChoiBMDSS17, DBLP:conf/nips/SrivastavaVRGS17, DBLP:journals/pvldb/ParkMGJPK18, DBLP:journals/corr/abs-1811-11264, DBLP:conf/nips/XuSCV19}, we extend the conditional tabular GAN (CTGAN)\,\cite{DBLP:conf/nips/XuSCV19}, which is the state-of-the-art method for generating realistic tabular data. CTGAN's key techniques are using mode-specific normalization to learn complicated column distributions and training-by-sampling to overcome imbalanced training data. 

\paragraph{Realistic Examples}

CTGAN does not guarantee that all constraints requiring domain knowledge are satisfied. For example, in the AdultCensus dataset, the marital status ``Wife'' means that the person is female, but we need to perform separate checking instead of relying on CTGAN. Our solution is to avoid certain patterns that are never seen in the original data. In our example, there are no examples where a Wife is a male, so we ignore all CTGAN-generated examples with this combination. This checking can be performed efficiently by identifying frequent feature pairs in the original data and rejecting any feature pair that does not appear in this list. In addition, we use clipping and quantization techniques to further make sure the feature values are valid.

\section{Experiments}
\label{sec:experiments}

\paragraph{Datasets} We use four real tabular datasets for binary and multi-class classification tasks. All the datasets contain people records whose information can be leaked. The last Diabetes dataset is large and thus used to demonstrate the scalability of our techniques.

\begin{itemize}
    \item AdultCensus\,\cite{adultcensus}: Contains 45,222 people examples and is used to determine if one has a salary of $\geq$\$50K per year. 
    \item COMPAS\,\cite{Compas}: Contains 7,214 examples and is used to predict criminal recidivism rates. 
    \item Epileptic Seizure Recognition (ESR)\,\cite{esr}: Contains 11,500 electroencephalographic (EEG) recording data and is used to classify five types of brain states including epileptic seizure.
    \item Diabetes\,\cite{Strack2014}: Contains 100,000 diabetes patient records in 130 US hospitals between 1999--2008. 
\end{itemize}

\paragraph{Target and Base Examples}

For each dataset, we choose 10 targets per dataset randomly. For each target, we choose $k$ nearest examples with different labels as the base examples to generate $k$ watermarked disinformation examples.

\paragraph{Measures}

To evaluate a PDB, we use {\em precision}, which is defined as the portion of examples that are on the other side of the decision boundary from the target and have different ground truth labels. To evaluate a model's performance, we measure the {\em accuracy}, which is the portion of predictions that are correct, and use the {\em confidence} given by the model. For all measures, we always report percentages.

\paragraph{Models}
We use three types of models: {\em surrogate labelers} for PDBs, {\em victim models} to simulate inaccessible black-box models, and {\em attack models} that are used to perform MIAs. 

We use 18 surrogate labelers explained below and summarized in Table~\ref{tbl:modelarchitectures}. Although we could use more complex models, they would overfit on our datasets.
\begin{itemize}
    \item Seven neural networks that have different combinations of the number of layers, the number of nodes per layer, and the activation function. We use the naming convention \model{s\_nn\_A\_X-Y}, which indicates a neural network that uses the activation function $A$ (\model{tanh}, \model{relu}, \model{log}, and \model{identity}) and has $X$ layers with $Y$ nodes per layer. 
    \item Two decision trees \model{s\_tree} and two random forests (\model{s\_rf}) using the Gini and Entropy purity measures.
    \item Four SVM models (\model{s\_svm}) using the radial basis function (\model{rbf}), linear, polynomial, and sigmoid kernels.
    \item Three other models: gradient boosting (\model{s\_gb}), AdaBoost (\model{s\_ada}), and logistic regression (\model{s\_logreg}).
\end{itemize}

For the victim models, we use a subset of Table~\ref{tbl:modelarchitectures} consisting of 13 models (four neural networks, four trees and forests, two SVMs, and three others), but with different numbers of layers and optimizers to clearly distinguish them from the surrogate labelers. For the attack models, we select nine of the smallest models having the fewest layers, depth, or number of tree estimators from Table~\ref{tbl:modelarchitectures}. We choose small models because attack models train on a victim model's output and loss and need to be small to perform well. We use the same naming conventions as Table~\ref{tbl:modelarchitectures} except that the model names start with ``a\_'' instead of ``s\_'' as shown in Table~\ref{tbl:mia}.

\begin{table}[t]
  \begin{tabular}{c|c}
    \toprule
    & Surrogate Labelers\\
    \midrule
 \multirow{2}{*}{\model{s\_nn}} &\model{relu\_5-2},\: \model{relu\_50-25},\:\model{relu\_200-100},\:\model{relu\_25-10} \\
 &\model{tanh\_5-2},\:\model{log\_5-2},\:\model{identity\_5-2} \\
 \midrule
 \multirow{1}{*}{\model{s\_tree}}&\model{dt\_gini},\:\model{dt\_entropy},\:\model{rf\_gini},\:\model{rf\_entropy} \\
\midrule
\multirow{1}{*}{\model{s\_svm}}&\model{rbf},\:\model{linear},\:\model{polynomial},\:\model{sigmoid}\\
\midrule
 \multirow{1}{*}{\model{others}}&\model{s\_gb},\:\model{s\_ada},\:\model{s\_logreg} \\
    \bottomrule
  \end{tabular}
  \caption{18 surrogate labeler architectures.}
  \label{tbl:modelarchitectures}
\end{table}

\paragraph{Methods}
We compare \method{} with three baselines: (1) {\em CPA} is the convex polytope attack extended to end-to-end training; (2) {\em GAN only} is \method{} using a CTGAN only; and (3) {\em WM only} is \method{} using watermarking only.

\paragraph{Other Settings}

We set the abstain threshold $\beta$ to 0.1. For all models, we set the learning rate to 1e-4 and the number of epochs to 1K. For CTGAN, we set the input random vector size to 100. We use PyTorch and Scikit-learn, and all experiments are performed using Nvidia Titan RTX GPUs. We evaluate all models on separate test sets. 


\subsection{Decision Boundary as a Labeler Proxy}
\label{sec:labelerproxy}

We evaluate the PDB precision in Table~\ref{tbl:precision}. We use cross validation accuracies to select the top-$k$ performing surrogate labelers without knowledge of the test accuracies (more details are in \ifthenelse{\boolean{techreport}}{the appendix}{our technical report\,\shortcite{redactortr}}). We then use the following groups of models as surrogate labelers: \model{g\_all} contains all the models, \model{g\_top-k} contains the top-performing surrogate labelers, \model{g\_nn\_only} contains the neural network models, \model{g\_tree\_only} contains the tree models, \model{g\_svm\_only} contains the SVM models, and \model{g\_others} contains the rest of the models. The table shows the PDB's precision for different $\alpha$ tolerance thresholds. 
For model groups with many surrogate labelers, the precision tends to increase for larger $\alpha$ values.

We observe that employing five or more surrogate labelers leads to good performance.
Compared to taking a majority vote of surrogate labelers (MV), the precision of a PDB is usually higher. In particular, using \model{g\_top-5} combined with $\alpha = 0.95$ results in the best precision, so we use this setup in the remaining sections.

\begin{table}[t]
  
  \begingroup
    \fontsize{9pt}{12pt}\selectfont
  \begin{tabular}{c|ccccc|c}
    \toprule
    Group & 0.5 & 0.7 & 0.9 & 0.95 & 0.99 & MV\\
    \midrule
     \model{g\_all} & 83.68 & 83.86 & 84.13 & 84.35 & 84.38 & 84.42\\
     \model{g\_top-15} & 84.52 & 84.72 & 85.11 & 85.22 & 85.59 & 84.37\\
     \model{g\_top-10} & 85.00 & 85.00 & 85.20 & 85.49 & \bf{86.28} & 83.90\\
     \model{g\_top-5} & 85.37 & 86.39 & \bf{87.95} & \bf{88.74} & 78.92 & 84.24\\
     \model{g\_top-3} & 85.30 & \bf{87.18} & 82.45 & 82.45 & 78.54 & 75.39\\
     \midrule
     \model{g\_nn-only} & 81.74 & 81.79 & 82.08 & 82.32 & 82.66 & 84.43\\
     \model{g\_tree-only} & \bf{85.44} & 86.12 & 87.86 & 88.18 & 79.34 & 84.35\\
     \model{g\_svm-only} & 82.96 & 82.96 & 82.96 & 82.96 & 64.20 & \bf{84.83}\\
     \model{g\_others} & 85.13 & 86.94 & 80.33 & 80.33 & 77.27 & 75.39\\
    \bottomrule
  \end{tabular}
  \endgroup
  \caption{Precision for PDBs with different $\alpha$ tolerance thresholds (0.5--0.99) and taking a majority vote of the surrogate labelers (MV).}
  \label{tbl:precision}
\end{table}

\subsection{Disinformation Performance}
\label{sec:disinfoperformance}

We evaluate \method{}'s disinformation in terms of how it changes a victim model's accuracy and confidence on the AdultCensus, COMPAS, and ESR in Table~\ref{tbl:confidenceaccuracy}. (\ifthenelse{\boolean{techreport}}{Evaluations on image data give similar results and are shown in the appendix.}{Evaluations on image data give similar results and are shown in our technical report\,\shortcite{redactortr}.})
We train the 13 victim models for each dataset. We then generate disinformation for the targets (500 examples for AdultCensus, 50 for COMPAS, and 100 for ESR) and re-train the victim models on the dataset plus disinformation. As a result, \method{} reduces the performances more than the other baselines (especially CPA) without reducing Test Acc. significantly.

\begin{table}[t]
  \setlength{\tabcolsep}{2pt}
  \begingroup
    \fontsize{9pt}{12pt}\selectfont
  \begin{tabular}{c|c|c|cc}
    \toprule
    & & Overall Test& Target Acc. & Target Conf.\\
     &\bf & \:Acc. Change\: & Change & Change \\
    \midrule
    \multirow{4}{*}{\makecell{Adult\\Census}}& {\em CPA} & -0.27$\pm$0.52 & -2.78$\pm$8.08 & -1.97$\pm$8.60 \\
    &{\em GAN only} & -0.49$\pm$0.65 & -16.67$\pm$13.72 & -13.30$\pm$10.69 \\
    &{\em WM only} &  -1.43$\pm$1.40 & -28.89$\pm$12.78 & -21.35$\pm$15.00 \\
    &\method{} & -1.99$\pm$1.73 & \bf -37.22$\pm$13.20 & \bf -26.23$\pm$14.44 \\
    \midrule
    \multirow{4}{*}{COMPAS}& {\em CPA} & -0.26$\pm$0.80 & -0.56$\pm$5.39 & -2.24$\pm$3.10 \\
    &{\em GAN only} & -0.14$\pm$0.72 & -5.56$\pm$10.96 & -2.30$\pm$3.31 \\
    &{\em WM only} & -2.31$\pm$2.08 & -32.77$\pm$20.23 & -21.68$\pm$13.26 \\
    &\method{} & -2.40$\pm$2.18 & \bf -33.89$\pm$18.83 & \bf -23.93$\pm$14.37 \\
    \midrule
    \multirow{4}{*}{ESR}& {\em CPA} & -0.43$\pm$4.27 & -7.14$\pm$16.04 & -2.59$\pm$5.75 \\
    &{\em GAN only} & -0.65$\pm$1.58 & -8.57$\pm$15.74 & -1.57$\pm$10.37 \\
    &{\em WM only} & -0.07$\pm$1.13 & -34.29$\pm$12.72 & -18.42$\pm$17.16 \\
    &\method{} & -0.11$\pm$0.89 & \bf -35.71$\pm$13.97 & \bf -18.28$\pm$17.25 \\
    \bottomrule
  \end{tabular}
  \endgroup
  \caption{Average performance change of victim models on targets when generating disinformation examples on the AdultCensus, COMPAS, and ESR datasets. The number of inserted examples is about 1\% of the entire dataset size.}
  \label{tbl:confidenceaccuracy}
\end{table}

\begin{figure}[t]
\centering
    \subfloat[]{
        {\includegraphics[scale=0.56]{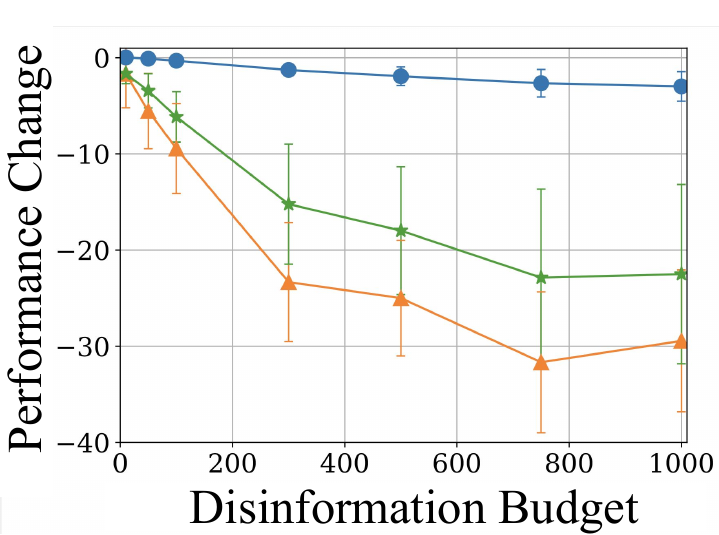}}
        \label{fig:disinformationcount}
        }
    \subfloat[]{
        {\includegraphics[scale=0.56]{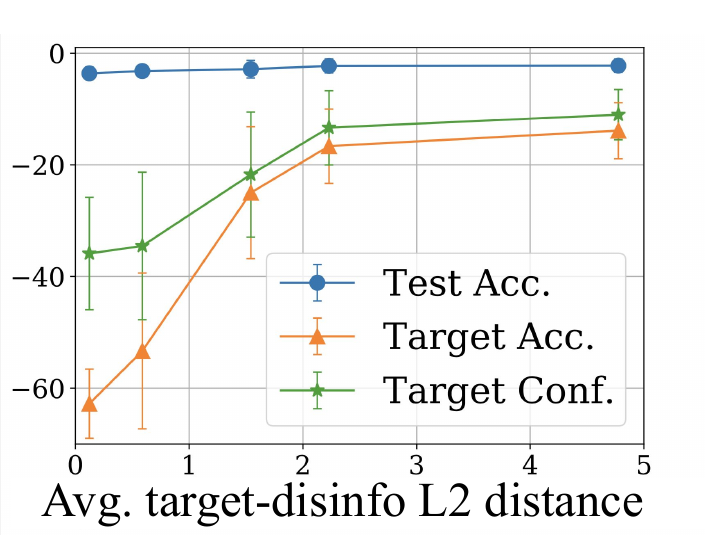}}
        \label{fig:vicinity}
        }
    \caption{(a) As the number of disinformation examples that we can create increases, the target accuracy and confidence decrease significantly while the overall test accuracy decreases only by 2\%. (b) As the distance to the target increases, we observe increasing trends as opposed to (a).}
\label{fig:trends}
\end{figure}

Using the same victim model setting, we also analyze how the disinformation budget and the distance between the target and disinformation impacts the disinformation performance. We first select 10 random target examples and vary the number of disinformation examples generated. Figure~\ref{fig:disinformationcount} shows how the average target accuracy and confidence of the 13 victim models decrease further as more disinformation budget is allowed, but eventually plateaus.
Next, we select 50 random target examples and generate disinformation. Then we cluster the targets by their average $L_2$ distances to their disinformation examples. We then plot each cluster in Figure~\ref{fig:vicinity}, which shows the average target accuracy and confidence degradation of the 13 models against the average target-disinformation $L_2$ distance of each cluster. As the disinformation is further away from a target, it becomes difficult to reduce the target's accuracy and confidence. 

\subsection{Defense Against Inference Attacks}
\label{sec:mia}

We evaluate \method{} against MIAs~\cite{shokri2017membership}. We use nine models in Table~\ref{tbl:modelarchitectures} with different hyperparameters for attacking the trained victim models.
We use the AdultCensus dataset and select 10 random target examples. Table~\ref{tbl:mia} shows the MIA performances with and without 200 disinformation examples using the nine attack models. For each scenario, we specify the attack model's overall $F_1$ score and average {\em target inference accuracy}, which is the fraction of target examples the attack model correctly predicts membership. We use the $F_1$ score just for this experiment to address the class imbalance of membership versus non-membership. Each experiment is repeated seven times. The less accurate the attack model, the better the privacy of the target. As a result, the overall $F_1$ score of the attack model does not change much, but the target accuracy decreases significantly (by up to 26.66\%) due to the disinformation. Furthermore some target accuracies drop to around 50\%, which means the classification is almost random. \ifthenelse{\boolean{techreport}}{Evaluations on other MIAs~\cite{lossmia,DBLP:conf/ndss/Salem0HBF019} give similar results and are shown in the appendix.}{Evaluations on other MIAs~\cite{lossmia,DBLP:conf/ndss/Salem0HBF019} give similar results and are shown in our technical report\,\shortcite{redactortr}.}

\begin{table}[t]
  \centering
  \setlength{\tabcolsep}{2pt}
  \begingroup
    \fontsize{9pt}{12pt}\selectfont
  \begin{tabular}{c|cc|cc|c}
    \toprule
    & \multicolumn{2}{c|}{Without Disinfo.} & \multicolumn{2}{c|}{With Disinfo.} & \\
    \toprule
    Attack & Overall & Target & Overall & Target & Target Acc.\\
    Model& F1 score & Acc. & F1 score & Acc. & Change\\
    \midrule
    \model{a\_tanh\_5-2} & 58.96 & 78.57 & 59.41 & 65.71 & -12.86 \\
    \model{a\_relu\_5-2} & 61.18 & 82.86 & 61.33 & 71.43 & -11.43 \\
    \model{a\_identity\_5-2} & 59.19 & 75.71 & 59.04 & 65.71 & -10.00 \\
    \model{a\_dt\_gini} & 52.02 & 73.33 & 51.04 & 46.67 & -26.66 \\
    \model{a\_dt\_entropy} & 52.22 & 60.00 & 51.66 & 43.33 & -16.67 \\
    \model{a\_rf\_gini} & 52.20 & 65.00 & 52.03 & 51.67 & -13.33 \\
    \model{a\_rf\_entropy} & 51.78 & 55.71 & 51.86 & 45.71 & -10.00 \\
    \model{a\_ada} & 53.85 & 61.43 & 52.74 & 54.29 & -7.14 \\
    \model{a\_logreg} & 61.30 & 80.00 & 61.20 & 70.00 & -10.00 \\
    \midrule
    Average & 55.86 & \bf 70.29 & 55.59 & \bf 57.17 & -13.12 \\
    \bottomrule
  \end{tabular}
  \endgroup
  \caption{Using \method{}'s disinformation to defend against MIAs using attack models. For 10 target examples, a total of 200 disinformation examples are generated. For each model, we show how the disinformation changes its performances.}
  \label{tbl:mia}
\end{table}

\subsection{Realistic Examples}

We perform a comparison of our disinformation with real data to see how realistic it is. 
Table~\ref{tbl:disinformation} shows a representative disinformation example (among many others) that was generated using our method along with the target and a few nearest neighbors.
To see if the disinformation is realistic, we conduct a poll asking 33 human experts to correctly identify five disinformation and five real examples. As a result, the average accuracy is 56.9\%, and the accuracies for identifying disinformation and real examples are 46.1\% and 67.8\%, respectively. We thus conclude that humans cannot easily distinguish our disinformation from real data, and that identifying disinformation is harder than identifying real data. 


\begin{table*}[t]
  \small
  \setlength{\tabcolsep}{2pt}
  
  \begingroup
    \fontsize{9pt}{12pt}\selectfont
  \begin{tabular}{c|cccccccccccc}
    \toprule
    &\bf Age& \bf Education & \bf Marital status & \bf Occupation & \bf Relationship & \bf Race & \bf Gender & \bf Capital gain & \bf Hrs/week & \bf Country & \bf Income \\
    \midrule
    \bf $T$ &38 & HS-grad & \makecell{Never-married} & \makecell{Machine-op-inspct} & Not-in-family & White & Male & 0 & 40 & US & $\leq$50K \\
    \midrule
    \bf $D$ & 43 & HS-grad & \makecell{Never-married} & \makecell{Machine-op-inspct} & Not-in-family & White & Male & 7676 & 40 & US & $>$50K \\
    \midrule
    \multirow{2}{*}{\bf $NN$} & 52 & HS-grad & \makecell{Never-married} & \makecell{Machine-op-inspct} & Not-in-family & White & Male & 0 & 45 & US & $>$50K \\
    & 36 & HS-grad & \makecell{Married-civ-spouse} & \makecell{Machine-op-inspct} & Husband & White & Male & 7298 & 40 & US & $>$50K \\

    \bottomrule
  \end{tabular}
  \endgroup
  
  \caption{Comparison of disinformation $D$ of target $T$ with $T$'s nearest neighbors $NN$ using the AdultCensus dataset.}
  \label{tbl:disinformation}
\end{table*}

\begin{table}[t]
  \setlength{\tabcolsep}{2pt}
  
  \begingroup
    \fontsize{9pt}{12pt}\selectfont
  \begin{tabular}{c|c|c|c|c|c|c}
    \toprule
    & \multicolumn{3}{c|}{Dist. Known} & \multicolumn{3}{c}{Dist. Unknown}\\
    \toprule
    $n$ & Time (s) & Local Acc. & $\Delta$ Acc. & Time (s) & Local Acc. & $\Delta$ Acc.\\
    \midrule
    1k & 59.72 & 58.82 & -30.00 & 47.35 & 66.64 & -31.67\\
    3k & 100.52 & 68.69 & -30.00 & 79.85 & 69.30 & -30.00\\
    5k & 125.76 & 70.40 & -31.67 & 109.28 & 69.84 & -27.77 \\
    7k & 162.78 & 72.30 & -26.67 & 122.68 & 70.42 & -20.37 \\
    \midrule
    All & 2,043.1 & 70.67 & -36.67 & 2,043.1 & 70.67 & -36.67 \\
    \bottomrule
  \end{tabular}
  \endgroup
  
  \caption{Evaluation of the two partial data strategies on the Diabetes dataset. For different data sizes ($n$), we show the average runtime for disinformation generation per target in seconds (Time), average local accuracy (Local Acc.), and average target accuracy change ($\Delta$ Acc.).}
  
  \label{tbl:scalability}
\end{table}

\subsection{Scalability}
\label{sec:scalability}

If the dataset is too large or not fully accessible, \method{} can still run on partial data. We evaluate \method{} on the Diabetes dataset by selecting 10 random targets, training the 18 surrogate labelers on nearest neighbors of the targets using Euclidean distance, and generating 200 disinformation examples per target. We use two strategies for selecting the partial data: (1) {\em Dist. Known}: We assume the entire class distribution is known and collect $n$ nearest neighbors of targets following this distribution (i.e., we effectively take a uniform sample of the entire data that is closest to the target) and (2) {\em Dist. Unknown}: We assume the distribution is unknown and collect nearest neighbors of the target until we have at least $n$ examples per class. 

In Table~\ref{tbl:scalability}, we compare the following for different $n$ values: (1) the average runtime for disinformation generation per target, (2) the average local accuracy, which is the average accuracy of surrogate labelers on the 10K nearest neighbors of each target, and (3) the average target accuracy change. As a result, when $n$ is at least 3,000 (3\% of the entire data), the runtime improves by $>$20x, while the average local accuracy and target accuracy change are comparable to the results using the entire data (All). In addition, utilizing the data distribution sometimes gives worse results than not due to the adjustment of class ratios of nearest neighbors to follow the entire distribution. Hence, using partial data without knowing the entire data distribution can be sufficient for effective disinformation.

\section{Related Work}
\label{sec:relatedwork}


\paragraph{Data Privacy, Data Deletion, and Disinformation}

Data privacy is a broad discipline of protecting one's personal information within data. The most popular approach is differential privacy\,\cite{DBLP:conf/tcc/DworkMNS06,DBLP:journals/cacm/Dwork11,DBLP:journals/fttcs/DworkR14} where random records are added to a database to lower the chance of information leakage. In comparison, we solve a subproblem of data privacy in ML where there is no control over the training data, and the only way to improve one's privacy is to add disinformation. 

Another related problem is data deletion where the goal is to make a model forget about certain data~\cite{DBLP:conf/nips/GinartGVZ19,DBLP:conf/icml/GuoGHM20,DBLP:conf/cvpr/GolatkarAS20,DBLP:conf/eccv/GolatkarAS20,DBLP:conf/aaai/GravesNG21}). Most of these techniques assume that the data or model can be changed at will. In comparison, we only assume that data can be added and that models may be trained with the new data at some point. 

Finally, disinformation has been studied in different contexts including data leakage detection\,\cite{DBLP:journals/tkde/PapadimitriouG11} and entity resolution\,\cite{DBLP:conf/cikm/WhangG13}. In comparison, \method{} focuses on obfuscating information in ML models for data privacy.

\paragraph{Targeted Poisoning}

Targeted poisoning attacks\,\cite{DBLP:conf/uss/SuciuMKDD18,DBLP:conf/nips/ShafahiHNSSDG18,DBLP:conf/icml/ZhuHLTSG19,DBLP:journals/titb/KermaniSRJ15} have the goal of flipping the predictions of specific targets to certain classes. Clean-label attacks\,\cite{DBLP:conf/uss/SuciuMKDD18,DBLP:conf/nips/ShafahiHNSSDG18,DBLP:conf/icml/ZhuHLTSG19} have been proposed for neural networks to alter the model's behavior on a specific test instance by poisoning the training set without having any control over the labeling. Convex Polytope Attack (CPA)~\cite{DBLP:conf/icml/ZhuHLTSG19} covers various structures of neural networks, which is different from  other techniques. All these techniques rely on a fixed feature space for optimization whereas \method{} does not assume this.


\section{Conclusion}
\label{sec:conclusion}

We proposed effective targeted disinformation methods for black-box models on structured data where there is no access to the labeling or model training. We explained why an end-to-end training setting is important and that existing targeted poisoning attacks that implicitly rely on a transferable learning setting do not perform well. We then presented \method{}, which is designed for end-to-end training where it generates a conservative probabilistic decision boundary to emulate labeling and then generates realistic disinformation examples that reduce the target's accuracy and confidence the most. Our experiments showed that \method{} generates disinformation more effectively than other targeted poisoning attacks, defends against MIAs, generates realistic disinformation, and scales to large data.

\section*{Acknowledgments}

This work was supported by the National Research Foundation of Korea(NRF) grant funded by the Korea government(MSIT) (No. NRF-2022R1A2C2004382) and by Samsung Electronics Co., Ltd. Steven E. Whang is the corresponding author.

\bibliography{main}

\clearpage

\ifthenelse{\boolean{techreport}}{

\appendix

\section{CPA Optimization in End-to-end Training}

Recall that we analyzed how CPA's optimization may not be effective in an end-to-end training scenario where the feature spaces is not fixed. To demonstrate this point, we run the extended version of CPA\,\cite{DBLP:conf/icml/ZhuHLTSG19} on the AdultCensus dataset using a neural network and generate a poison example $p$. We then observe how the relative $L_2$ distances between $p$ and $t$ change (Figure~\ref{fig:convexpolytope}). As the model trains in a transfer learning scenario, the feature distance from $p$ to $t$ decreases on three different layers in the model (dotted lines). However, when the model trains end-to-end, the feature distance from $p$ to $t$ increases rapidly (solid lines), which means that the model no longer makes the same classification for $p$ and $t$. 

\begin{figure}[h]
\centering
    \includegraphics[scale=0.38]{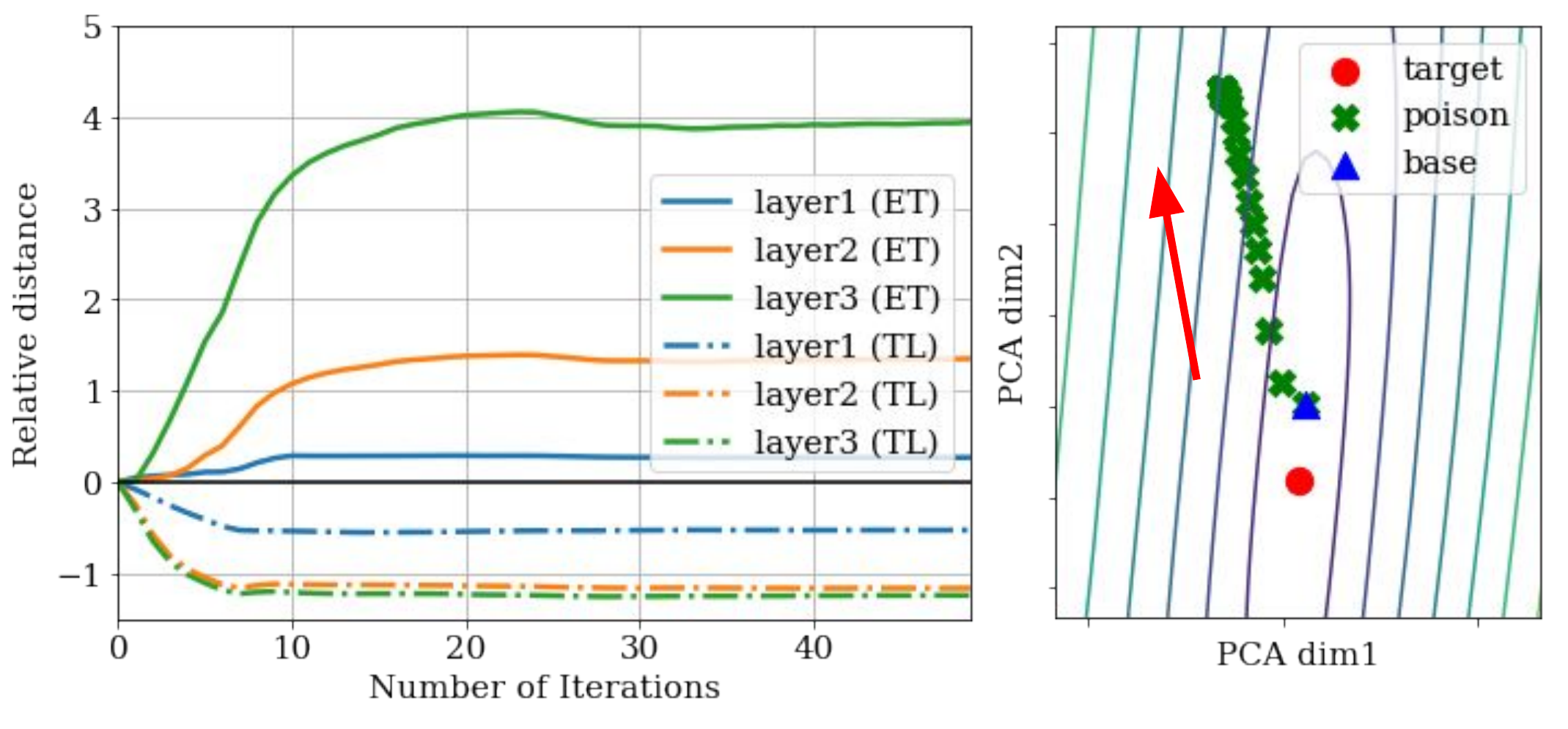}
    \caption{(Left) We run CPA\,\cite{DBLP:conf/icml/ZhuHLTSG19} on the AdultCensus dataset and observe the relative $L_2$ distances from the initial points depicted in Figure~\ref{fig:scenarios}. We measure distances from the poison example $p$ to target $t$ on possible feature spaces (different layers of a neural network). The transfer learning results (dotted lines) show how CPA is effective in reducing the distances from $p$ to $t$, while the end-to-end training results (solid lines) show how it fails to do so. (Right) The target, poison, and base points on one of the feature spaces using PCA~\cite{pca} for dimensionality reduction.}
    \label{fig:convexpolytope}
\end{figure}

\section{Algorithm}
Algorithm~\ref{alg:algorithm} shows the overall algorithm of \method{}. We first select random base examples that are preferably close to the target, but obviously have different labels according to our judgement (Step 2). We then generate candidate disinformation examples using watermarking and a CTGAN (Steps 3--9). We also construct the probabilistic decision boundary by combining good-performing surrogate models into a probabilistic model (Steps 10--15). Finally, we return the disinformation examples that are on the other side of the decision boundary from the target (Steps 16--22).


\begin{algorithm}[tb]
\caption{Pseudo code for generating disinformation.}
\label{alg:algorithm}
\textbf{Input}: Target example $t$, available data $I$, trained surrogate labelers $\phi$, trained generator model $G$, number of disinformation examples $N_d$, number of generated samples $N_{gen}$, tolerance threshold $\alpha$, abstain threshold $\beta$\\
\textbf{Output}: Disinformation examples $R$ \\
\begin{algorithmic}[1] 
    \STATE $C_{real} \leftarrow [\,]$ 
    \STATE $B \leftarrow $ NearestExamples($I$, $t$, $N_d$) $\:\:s.t.\:\:c_b\neq c_t$
    \FOR {$i$ in 0...$r$}
        \STATE $\gamma \leftarrow i/r$ 
        \STATE $C_{cand}$.append( WaterMarking($B$, $t$, $\gamma$) )
    \ENDFOR
    \STATE $C_{GAN} \leftarrow$ G.generate($N_{gen}$*$n$)
    \STATE $C_{GAN}\leftarrow$ FilterUnrealisticRecord($C_{GAN}$, $I$)
    \STATE $C_{real}$.append( NearestExamples($C_{GAN}$, $t$, $N_{gen}$) )\
    
    \STATE $\phi_{topK} \leftarrow$ SelectTopKmodels($I$, $\phi$, $k$)
    \FOR{$\phi_i\in \phi$}
        \STATE $\Phi_i \leftarrow \phi_i(I)$ 
    \ENDFOR
    \STATE $\Lambda \leftarrow $ LabelMatrixTransform($\Phi$, $\beta$)
    
    \STATE $M \leftarrow $ TrainGenModel($\Lambda$, $I$)
    \STATE $R \leftarrow [\,]$
    \FOR{$j$ in 1...$N_d$}
        \STATE $d_j \leftarrow \argmin_{x\in C_{cand}} ||x-t||^2\:\:s.t. \:\:M_c(x)\geq\alpha$ 
        \STATE $R$.append($d_j$)
        \STATE $C_{cand}$.remove($d_j$)
    \ENDFOR
    \STATE \textbf{return} $R$
\end{algorithmic}
\end{algorithm}

\section{Performance of Surrogate Labelers}

Recall that we use Cross Validation (CV) accuracies to select the top-$k$ performing surrogate labelers without knowledge of the test accuracies. In Table~\ref{tbl:modelperformances}, we show that the surrogate models' Train, Test, and CV accuracies on the AdultCensus dataset. The table shows that it is reasonable to select the top-$k$ models using CV accuracies since those are similar to test accuracies.

\begin{table}[t]
  \centering
  \begingroup
    \fontsize{9pt}{12pt}\selectfont
  \begin{tabular}{c|c|ccc}
    \toprule
    & Surrogate Labeler &Train Acc.&Test Acc.&CV Acc.\\
    \midrule
 \multirow{8}{*}{\model{s\_nn}}&\model{tanh\_5-2} & 86.46 & 85.07 & 84.24 \\
 &\model{relu\_5-2} & 86.57 & 85.24 & 84.92 \\
 &\model{relu\_50-25} & 90.33 & 82.44 & 82.67 \\
 &\model{relu\_200-100} & 95.55 & 81.56 & 81.64 \\
 &\model{relu\_25-10} & 87.93 & 84.22 & 83.64 \\
 &\model{log\_5-2} & 85.63 & 85.26 & 84.50\\
 &\model{identity\_5-2} & 84.84 & 84.74 & 84.79\\
 \midrule
 \multirow{4}{*}{\model{s\_tree}}&\model{dt\_gini} & 85.28 & 85.56 & 84.73 \\
 &\model{dt\_entropy} & 85.29 & 85.38 & 84.84 \\
 &\model{rf\_gini} & 85.08 & 85.21 & 84.93 \\
 &\model{rf\_entropy} & 85.16 & 85.37 & 84.96 \\
\midrule
\multirow{4}{*}{\model{s\_svm}}&\model{rbf} & 85.83 & 84.92 & 84.53 \\
 &\model{linear} & 84.78 & 85.03 & 84.63 \\
 &\model{polynomial} & 85.13 & 83.16 & 82.79 \\
 &\model{sigmoid} & 81.24 & 82.11 & 82.22 \\
\midrule
 \multirow{3}{*}{\model{others}}&\model{s\_gb} & 85.70 & 85.99 & 86.12 \\
 &\model{s\_ada} & 86.22 & 86.30 & 86.12 \\
 &\model{s\_logreg} & 84.90 & 84.86 & 84.76  \\
    \bottomrule
  \end{tabular}
  \endgroup
  \caption{18 surrogate labelers' model performances.}
  \label{tbl:modelperformances}
\end{table}

\section{Disinformation Performance on Image Data}

Our techniques can be extended to images, but the key issue is whether transfer learning is used. If so, the feature space is fixed and can be utilized as in CPA. Our method is most effective when transfer learning is not possible. To demonstrate this point, we perform an experiment comparing \method{} and CPA on the MNIST dataset. This dataset does not need transfer learning as it has relatively smaller images than other datasets that are easier to classify without having to use pre-trained weights. As a result, \method{} performs better than CPA as shown in Table~\ref{tbl:image}.
We also note that \method{} need to be updated to fully support image data. For example, the clipping and quantization techniques cannot be utilized for image data, and CTGAN is used for generating tabular data. We can replace these techniques with corresponding techniques for image data.

\begin{table}[t]
  \setlength{\tabcolsep}{2pt}
  \begingroup
    \fontsize{9pt}{12pt}\selectfont
  \begin{tabular}{c|c|c|cc}
    \toprule
    & & Overall Test& Target Acc. & Target Conf.\\
     &\bf & \:Acc. Change\: & Change & Change \\
    \midrule
    \multirow{2}{*}{MNIST}& {\em CPA} & -2.45$\pm$3.64 & -8.33$\pm$13.29 & -1.78$\pm$3.61 \\
    &\method{} & -3.51$\pm$3.74 & \bf -15.00$\pm$17.61 & \bf -6.17$\pm$8.08 \\
    \bottomrule
  \end{tabular}
  \endgroup
  \caption{Average performance change of victim models on targets when generating disinformation examples on the MNIST dataset. The number of inserted examples is about 1\% of the entire dataset size.}
  \label{tbl:image}
\end{table}

\section{Evaluation of \method{} against Other MIAs}

Recall that we empirically showed how \method{} is effective in defending against a representative MIA\,\cite{shokri2017membership}. In Table~\ref{tbl:mia2}, we show that \method{} is also effective against two other popular MIAs: Loss MIA~\cite{lossmia} and Conf. MIA~\cite{DBLP:conf/ndss/Salem0HBF019}, which do not require the attack models, but only threshold values on the loss and confidence score, respectively. We use ROC AUC to measure the average performance over changing thresholds. As a result, we observe similar results as Table~\ref{tbl:mia} where the overall AUC of the attack model does not change much, but the target AUC decreases significantly.

\begin{table}[t]
  \setlength{\tabcolsep}{2pt}
  \begin{tabular}{c|cc|cc|c}
    \toprule
    & \multicolumn{2}{c|}{Without Disinfo.} & \multicolumn{2}{c|}{With Disinfo.} & \\
    \toprule
    Threshold  & Overall & Target & Overall & Target & Target AUC\\
    MIA Type & AUC & AUC & AUC & AUC & Change\\
    \midrule
    Loss MIA& 49.93 & 88.75 & 49.85 & 63.96 & -24.79\\
    \midrule
    Conf. MIA& 50.55 & 88.75 & 50.52 & 69.79 & -18.96 \\
    \bottomrule
  \end{tabular}
  \caption{Using \method{}'s disinformation to defend against MIAs that do not use attack models. The other conditions are identical to Table~\ref{tbl:mia}.}
  \label{tbl:mia2}
\end{table}

}

\end{document}

%% file: math_commands.tex
\usepackage{amsmath,amsfonts,bm}

\def\eqref#1{equation~\ref{#1}}

\def\1{\bm{1}}

\DeclareMathAlphabet{\mathsfit}{\encodingdefault}{\sfdefault}{m}{sl}
\SetMathAlphabet{\mathsfit}{bold}{\encodingdefault}{\sfdefault}{bx}{n}

\DeclareMathOperator*{\argmax}{arg\,max}
\DeclareMathOperator*{\argmin}{arg\,min}